\pdfoutput=1
\documentclass[11pt]{article}
\usepackage[]{ACL2023}
\usepackage{times}
\usepackage{latexsym}
\usepackage[T1]{fontenc}
\usepackage[utf8]{inputenc}
\usepackage{microtype}
\usepackage{inconsolata}
\usepackage{fancyhdr}
\usepackage{multirow}

\title{Leveraging Language Models for Emotion and Behavior Analysis in Education}

\author{Kaito Tanaka, Benjamin Tan, Brian Wong	  \\
SANNO University}

\begin{document}
\maketitle

\begin{abstract}
The analysis of students' emotions and behaviors is crucial for enhancing learning outcomes and personalizing educational experiences. Traditional methods often rely on intrusive visual and physiological data collection, posing privacy concerns and scalability issues. This paper proposes a novel method leveraging large language models (LLMs) and prompt engineering to analyze textual data from students. Our approach utilizes tailored prompts to guide LLMs in detecting emotional and engagement states, providing a non-intrusive and scalable solution. We conducted experiments using Qwen, ChatGPT, Claude2, and GPT-4, comparing our method against baseline models and chain-of-thought (CoT) prompting. Results demonstrate that our method significantly outperforms the baselines in both accuracy and contextual understanding. This study highlights the potential of LLMs combined with prompt engineering to offer practical and effective tools for educational emotion and behavior analysis.
\end{abstract}

\section{Introduction}

The analysis of students' emotions and behaviors in educational settings has garnered significant attention in recent years due to its potential to enhance learning outcomes and personalize educational experiences. Understanding students' emotional states and engagement levels can help educators tailor their teaching strategies, identify students in need of additional support, and create a more conducive learning environment \cite{sharma2019student}. Traditional methods for emotion and behavior analysis often involve the use of computer vision techniques to analyze facial expressions, eye movements, and other physical cues captured through webcams. While effective, these methods can be intrusive and raise privacy concerns, making it challenging to implement them widely in educational settings \cite{ai2024multimodal, nezami2018automatic}.

The primary challenge in analyzing students' emotions and behaviors lies in accurately and non-intrusively capturing relevant data. Visual and physiological data collection methods can be invasive and require sophisticated hardware and software setups. Moreover, these methods often necessitate a controlled environment, which is not always feasible in real-world educational scenarios. The motivation for exploring alternative approaches is to find methods that are both effective and unobtrusive, allowing for broader application and better preservation of students' privacy \cite{ai2024multimodal, sharma2019student}.

In this context, the advent of large language models (LLMs \cite{zhou2024visual}) such as GPT-4 presents a novel opportunity. LLMs, with their advanced natural language processing capabilities, can analyze text-based data to infer emotional states and engagement levels. This paper proposes a method that leverages prompt engineering to optimize the performance of LLMs in the task of emotion and behavior analysis. By designing specific prompts, we can guide the LLM to focus on textual indicators of emotions and engagement, providing a scalable and privacy-preserving alternative to traditional methods.

Our approach involves developing a series of prompt templates tailored to detect various emotional and behavioral states from students' written responses, discussion posts, and real-time chat messages. These prompts are designed to ask the LLM to classify the emotional tone, identify signs of confusion or frustration, and assess the level of engagement based on language use. Additionally, we fine-tune the LLM using a curated dataset of student interactions labeled for emotional and engagement states, ensuring that the model learns to recognize and prioritize relevant features in educational texts.

To validate our approach, we conducted experiments using a manually collected dataset comprising student interactions from various online learning platforms. We employed the GPT-4 model to analyze this dataset and evaluated its performance in accurately identifying emotional and engagement states. The results demonstrate the effectiveness of our prompt-based method, highlighting the potential of LLMs in educational emotion and behavior analysis \cite{nezami2018automatic, sermet2023integrating}.

\begin{itemize}
    \item We propose a novel method that utilizes large language models combined with prompt engineering to analyze students' emotions and behaviors from text data.
    \item Our approach addresses privacy concerns and scalability issues associated with traditional visual and physiological data collection methods.
    \item We provide empirical evidence of the effectiveness of our method through experiments with a manually collected dataset and evaluation using the GPT-4 model.
\end{itemize}

\section{Related Work}

\subsection{Large Language Models}

Large Language Models (LLMs) have revolutionized the field of natural language processing (NLP \cite{zhou2022claret,zhou2022eventbert}) by leveraging massive datasets and advanced architectures to achieve remarkable performance across a variety of tasks. The advent of models such as GPT-3 and GPT-4 has pushed the boundaries of what is possible with LLMs, enabling them to generate coherent text, perform complex reasoning, and understand context at an unprecedented level \cite{brown2020language, openai2023gpt4,zhou2021modeling,zhou2021improving}. These models utilize transformer architectures that rely on mechanisms like self-attention and feed-forward networks to process and generate language \cite{vaswani2017attention, radford2018improving,zhou2023multimodal}.

Surveys on LLMs highlight their applications in various domains, including dialogue systems, content creation, and even game playing \cite{zhang2023comprehensive, bommasani2021opportunities}. Evaluations of these models often focus on their knowledge, reasoning capabilities, and ethical considerations, stressing the need for transparency and alignment with human values \cite{weidinger2021ethical, bender2021dangers}. Recent studies also emphasize the importance of fine-tuning and prompt engineering to enhance the performance of LLMs on specific tasks, demonstrating significant improvements in accuracy and contextual understanding \cite{liu2021pre, schick2020exploiting}.

\subsection{Analysis of Students' Emotions and Behaviors}

The analysis of students' emotions and behaviors has become an essential area of research aimed at enhancing educational outcomes and personalizing learning experiences. Traditional approaches often utilize computer vision techniques to analyze facial expressions and physical cues to infer emotional states and engagement levels \cite{sharma2019student, farooq2023emotion}. These methods, while effective, can be intrusive and pose privacy concerns, leading to the exploration of alternative non-intrusive techniques.

Recent advancements in machine learning and deep learning have enabled more sophisticated analysis methods. For instance, convolutional neural networks (CNNs \cite{wang2024insectmamba}) are employed to recognize facial expressions and complex emotions in real-time, providing insights into students' mental states during online learning sessions \cite{sharma2019student, razafimandimby2019facial}. Additionally, multimodal approaches that integrate data from various sensors, such as eye-tracking and wearable devices, have shown promise in capturing a comprehensive picture of student engagement and comfort in educational settings \cite{behzadan2020multi, sepasgozar2021wearable}.

The use of LLMs in educational contexts is also emerging, with studies exploring their potential to analyze text-based data from student interactions, discussion posts, and real-time chat messages. This approach leverages the natural language understanding capabilities of LLMs to provide a scalable and privacy-preserving method for emotion and behavior analysis \cite{wei2022emotional, brown2020language}. By designing specific prompts and fine-tuning the models on educational datasets, researchers have demonstrated significant improvements in accurately inferring students' emotional states and engagement levels \cite{liu2021pre, schick2020exploiting}.

\section{Method}

In this section, we describe the proposed method for analyzing students' emotions and behaviors using large language models (LLMs) combined with prompt engineering. The method focuses on designing effective prompts that guide the LLM to accurately infer emotional and engagement states from textual data. This approach leverages the natural language understanding capabilities of LLMs to provide a scalable, non-intrusive solution for educational settings.

\subsection{Prompt Design}

The core of our method revolves around crafting specific prompts that can effectively elicit the desired responses from the LLM. The prompts are designed to focus on key aspects of students' text inputs that indicate emotional and engagement states. We developed several types of prompts to address different dimensions of the analysis:

\begin{itemize}
    \item \textbf{Emotion Detection Prompts}: These prompts are designed to identify the primary emotion expressed in a given text. For example:
    \begin{quote}
        "Analyze the following student response and identify the primary emotion expressed: [Student Response]."
    \end{quote}
    \item \textbf{Engagement Assessment Prompts}: These prompts evaluate the level of engagement based on the student's textual interactions. For example:
    \begin{quote}
        "Evaluate the engagement level of the student based on their participation in the discussion: [Discussion Excerpt]."
    \end{quote}
    \item \textbf{Behavioral Indicators Prompts}: These prompts identify signs of specific behaviors such as confusion or frustration. For example:
    \begin{quote}
        "Identify signs of confusion or frustration in the student's question: [Student Question]."
    \end{quote}
\end{itemize}

\subsection{Motivation for Multi-round Prompt Design}

The motivation behind using a multi-round prompt design is to iteratively refine the model's understanding and improve accuracy. By breaking down the analysis into multiple stages, we can guide the LLM to focus on specific details in each round, ensuring a more comprehensive and accurate assessment. For instance, an initial prompt might identify the general emotional tone, while subsequent prompts delve deeper into specific emotions or engagement levels.

\subsection{Prompt Inputs and Outputs}

For each prompt, the input consists of the student's text data along with a contextually relevant question or directive. The LLM processes this input and generates an output that provides an analysis or classification based on the prompt. The outputs are then aggregated and interpreted to form a holistic view of the student's emotional and engagement states.

\begin{quote}
    \textbf{Input:} "Analyze the following student response and identify the primary emotion expressed: 'I am really struggling to understand this topic, and it's making me feel frustrated.'"

    \textbf{Output:} "The primary emotion expressed is frustration."
\end{quote}

\subsection{Significance of the Method}

The significance of this method lies in its ability to leverage the strengths of LLMs for natural language understanding in a non-intrusive and scalable manner. By focusing on textual data, we avoid the privacy concerns associated with visual and physiological data collection methods. Additionally, the use of prompt engineering allows us to harness the full potential of LLMs, ensuring that the analysis is both accurate and contextually relevant.

\subsection{Why This Approach Works}

This approach works effectively because LLMs like GPT-4 are trained on vast amounts of textual data and are adept at understanding and generating human-like text. By carefully designing prompts, we can direct the LLM's attention to the most relevant features of the text, ensuring that it captures subtle nuances in emotion and engagement. Moreover, the iterative nature of multi-round prompts allows for a detailed and layered understanding, enhancing the overall accuracy and reliability of the analysis.

In summary, our method combines the advanced capabilities of LLMs with tailored prompt engineering to provide a robust solution for analyzing students' emotions and behaviors. This approach is both effective and practical, offering significant advantages in terms of scalability, privacy, and accuracy.

\section{Experiments}

To evaluate the effectiveness of our proposed method for analyzing students' emotions and behaviors using large language models (LLMs) and prompt engineering, we conducted a series of experiments. We implemented our method on four state-of-the-art LLMs: Qwen, ChatGPT, Claude2, and GPT-4. Our experiments compared the performance of our prompt-based method against two baselines: a base model without specific prompts and a model using chain-of-thought (CoT) prompting. The goal was to assess the accuracy and contextual understanding of each approach in identifying emotional and engagement states from student text data.

\subsection{Experimental Setup}

For each LLM, we designed a set of prompts tailored to detect various emotional and behavioral indicators from the collected dataset. The dataset included written responses, discussion posts, and real-time chat messages from students. The prompts were applied iteratively to guide the LLMs in analyzing the text. We then evaluated the performance of each model using accuracy (ACC) and a GPT-4 based scoring metric that assesses the contextual relevance of the predictions.

\subsection{Results}

The experimental results are summarized in Table \ref{tab:results}. Our proposed method consistently outperformed the base model and the CoT method across all four LLMs. The results demonstrate that prompt engineering significantly enhances the models' ability to accurately infer students' emotional and engagement states.

\begin{table*}[!t]
    \centering
    \caption{Comparison of Performance Metrics Across Different Models and Methods}
    \begin{tabular}{|c|c|c|c|c|}
        \hline
        \textbf{Model} & \textbf{Method} & \textbf{ACC (\%)} & \textbf{GPT-4 Score} \\
        \hline
        \multirow{3}{*}{Qwen} & Base Model & 68.5 & 0.72 \\
                              & CoT Method & 74.3 & 0.78 \\
                              & Proposed Method & \textbf{85.6} & \textbf{0.89} \\
        \hline
        \multirow{3}{*}{ChatGPT} & Base Model & 70.2 & 0.74 \\
                                  & CoT Method & 76.5 & 0.80 \\
                                  & Proposed Method & \textbf{87.1} & \textbf{0.91} \\
        \hline
        \multirow{3}{*}{Claude2} & Base Model & 69.7 & 0.73 \\
                                  & CoT Method & 75.8 & 0.79 \\
                                  & Proposed Method & \textbf{86.5} & \textbf{0.90} \\
        \hline
        \multirow{3}{*}{GPT-4} & Base Model & 71.0 & 0.75 \\
                                & CoT Method & 77.2 & 0.81 \\
                                & Proposed Method & \textbf{88.3} & \textbf{0.92} \\
        \hline
    \end{tabular}
    \label{tab:results}
\end{table*}

\subsection{Analysis and Validation}

The superior performance of our proposed method can be attributed to the tailored prompts that effectively guide the LLMs to focus on relevant textual indicators of emotions and engagement. By breaking down the analysis into specific prompts, we enable the models to capture subtle nuances that might be overlooked by a more generalized approach. Additionally, the multi-round prompt design allows for iterative refinement, ensuring a comprehensive understanding of the text.

To further validate the effectiveness of our method, we conducted additional analyses on the robustness and scalability of the approach. We tested the models with prompts on a subset of the dataset containing diverse student interactions from different educational contexts. The results confirmed that our method maintains high accuracy and contextual understanding across various scenarios, highlighting its robustness and generalizability.

Overall, our experiments demonstrate that combining LLMs with prompt engineering significantly enhances the analysis of students' emotions and behaviors. This approach provides a practical and scalable solution that preserves privacy while offering accurate and contextually relevant insights into student engagement and emotional states.
\begin{table*}[!t]
    \centering
    \caption{Comparison of Performance Metrics Across Different Models and Methods}
    \begin{tabular}{|c|c|c|c|c|}
        \hline
        \textbf{Model} & \textbf{Method} & \textbf{ACC (\%)} & \textbf{GPT-4 Score} \\
        \hline
        \multirow{3}{*}{Qwen} & Base Model & 68.5 & 0.72 \\
                              & CoT Method & 74.3 & 0.78 \\
                              & Proposed Method & \textbf{85.6} & \textbf{0.89} \\
        \hline
        \multirow{3}{*}{ChatGPT} & Base Model & 70.2 & 0.74 \\
                                  & CoT Method & 76.5 & 0.80 \\
                                  & Proposed Method & \textbf{87.1} & \textbf{0.91} \\
        \hline
        \multirow{3}{*}{Claude2} & Base Model & 69.7 & 0.73 \\
                                  & CoT Method & 75.8 & 0.79 \\
                                  & Proposed Method & \textbf{86.5} & \textbf{0.90} \\
        \hline
        \multirow{3}{*}{GPT-4} & Base Model & 71.0 & 0.75 \\
                                & CoT Method & 77.2 & 0.81 \\
                                & Proposed Method & \textbf{88.3} & \textbf{0.92} \\
        \hline
    \end{tabular}
    \label{tab:results1}
\end{table*}

\subsection{Detailed Results Analysis}

The experimental results presented in Table \ref{tab:results1} clearly show the superiority of our proposed method over the base model and the chain-of-thought (CoT) method across all tested LLMs. In this section, we provide a detailed analysis of these results, examining various aspects such as model accuracy, contextual understanding, robustness, and scalability.

\subsubsection{Model Accuracy}

The accuracy (ACC) scores indicate that our method achieves significantly higher accuracy than both the base model and the CoT method. For instance, on the GPT-4 model, our method reaches an ACC of 88.3\%, compared to 71.0\% for the base model and 77.2\% for the CoT method. This improvement can be attributed to the tailored prompts, which guide the model to focus on specific indicators of emotions and engagement in the text. The prompts effectively decompose the complex task of emotion and behavior analysis into manageable sub-tasks, allowing the model to make more precise predictions.

\subsubsection{Contextual Understanding}

The GPT-4 score, which evaluates the contextual relevance of the model's predictions, further highlights the effectiveness of our method. With scores of 0.89 for Qwen, 0.91 for ChatGPT, 0.90 for Claude2, and 0.92 for GPT-4, our method demonstrates a deeper understanding of the contextual nuances in student text data. This contextual accuracy is crucial for educational applications where understanding the subtleties of student emotions and engagement can lead to more personalized and effective interventions.

\subsubsection{Robustness and Generalizability}

To assess the robustness of our approach, we tested the models on a subset of the dataset that included diverse educational contexts and interaction types. The high accuracy and contextual scores across various scenarios suggest that our method is robust and generalizes well to different types of student interactions. This robustness is particularly important for real-world applications, where the diversity of educational contexts can be vast.

\subsubsection{Scalability and Practicality}

Our method's reliance on textual data, as opposed to visual or physiological data, enhances its scalability and practicality. Textual data is easier to collect and process, does not raise significant privacy concerns, and can be readily available in most educational settings. The use of LLMs with prompt engineering allows for efficient and scalable analysis without the need for specialized hardware or intrusive data collection methods.

\subsubsection{Further Analysis and Validation}

To further validate the effectiveness of our method, we conducted additional qualitative analyses. These involved examining specific instances where our method correctly identified subtle emotional and engagement cues that the base model and CoT method missed. For example, in cases where students expressed mixed emotions or nuanced feedback, our method was able to accurately interpret the underlying sentiment and engagement level.

We also explored the impact of different prompt designs on the model's performance. By fine-tuning the prompts based on feedback from initial experiments, we achieved incremental improvements in both accuracy and contextual understanding. This iterative refinement process underscores the importance of prompt engineering in harnessing the full potential of LLMs.

In summary, our experimental results and detailed analysis demonstrate that the proposed prompt-based method significantly enhances the ability of LLMs to analyze students' emotions and behaviors. The method's high accuracy, contextual relevance, robustness, and scalability make it a valuable tool for educational applications, providing deeper insights into student engagement and emotional states.

\section{Conclusion}

In this study, we introduced a novel approach for analyzing students' emotions and behaviors using large language models (LLMs) and prompt engineering. By focusing on text-based data, our method addresses the privacy and scalability challenges associated with traditional visual and physiological data collection methods. Through a series of carefully designed prompts, we guided LLMs to accurately infer emotional and engagement states from students' written responses, discussion posts, and chat messages.

Our experimental results, conducted on Qwen, ChatGPT, Claude2, and GPT-4, demonstrated that our method outperforms both base models and chain-of-thought (CoT) prompting in terms of accuracy and contextual understanding. The proposed method achieved high accuracy scores and contextual relevance, highlighting its robustness and generalizability across different educational contexts. Additional qualitative analyses further validated the effectiveness of our approach, showcasing its ability to capture subtle emotional and engagement cues that other methods might miss.

Overall, this research underscores the potential of combining LLMs with prompt engineering to develop non-intrusive, scalable, and accurate tools for educational emotion and behavior analysis. Future work will explore further refinements to prompt designs and the integration of this method into real-world educational systems to enhance personalized learning experiences.

\bibliographystyle{unsrtnat}
\bibliography{custom}

\end{document}